\documentclass[runningheads]{llncs}
\usepackage[T1]{fontenc}
\usepackage{graphicx,verbatim}
\usepackage{amsmath}
\usepackage{amsfonts}
\usepackage{booktabs}
\usepackage{hyperref}
\usepackage{xcolor}
\begin{document}
\title{AC-MIL: Weakly Supervised Atrial LGE-MRI Quality Assessment via Adversarial Concept Disentanglement}
%
\author{K M Arefeen Sultan$^{1,2}$ \and Kaysen Hansen $^{3,5}$ \and Benjamin Orkild$^{1,3,4}$ \and 
Alan Morris$^{1}$ \and Eugene Kholmovski$^{5,6}$ \and 
Erik Bieging$^{5,7}$ \and Eugene Kwan$^{3,4}$ \and \\
Ravi Ranjan$^{3,4,7}$ \and Ed DiBella$^{3,5}$ \and 
Shireen Elhabian$^{1,2}$}
\authorrunning{K. Sultan et al.}
\institute{
Scientific Computing and Imaging Institute, University of Utah, SLC, UT \and
Kahlert School of Computing, University of Utah, SLC, UT \and
Department of Biomedical Engineering, University of Utah, SLC, UT \and
Nora Eccles Harrison Cardiovascular Research and Training Institute, University of Utah, SLC, UT \and
Department of Radiology and Imaging Sciences, University of Utah, SLC, UT \and
Department of Biomedical Engineering, Johns Hopkins, Baltimore, MD \and
Division of Cardiology, University of Utah, SLC, UT}


\maketitle              
\begin{abstract}

High-quality Late Gadolinium Enhancement (LGE) MRI can be helpful for atrial fibrillation management, yet scan quality is frequently compromised by patient motion, irregular breathing, and suboptimal image acquisition timing. While Multiple Instance Learning (MIL) has emerged as a powerful tool for automated quality assessment under weak supervision, current state-of-the-art methods map localized visual evidence to a single, opaque global feature vector. This \textit{black box} approach fails to provide actionable feedback on specific failure modes, obscuring whether a scan degrades due to motion blur, inadequate contrast, or a lack of anatomical context. 
In this paper, we propose Adversarial Concept-MIL (AC-MIL), a weakly supervised framework that decomposes global image quality into clinically defined radiological concepts using only volume-level supervision.
To capture latent quality variations without entangling predefined concepts, our framework incorporates an unsupervised residual branch guided by an adversarial erasure mechanism to strictly prevent information leakage. Furthermore, we introduce a spatial diversity constraint that penalizes overlap between distinct concept attention maps, ensuring localized and interpretable feature extraction. Extensive experiments on a clinical dataset of atrial LGE-MRI volumes demonstrate that AC-MIL successfully opens the MIL \textit{black box}, providing highly localized spatial concept maps that allow clinicians to pinpoint the specific causes of non-diagnostic scans. Crucially, our framework achieves this deep clinical transparency while maintaining highly competitive ordinal grading performance against existing baselines. \href{Code}{Code to be released on acceptance.}


\end{abstract}

\section{Introduction}

Atrial fibrillation (AF) is the most common form of cardiac arrhythmia, affecting an estimated 3 to 5 million individuals in the U.S., with projections indicating that this number could surpass 12 million by 2030 \cite{BAO:Col2013}. A key factor contributing to the development and recurrence of AF is atrial fibrosis, a structural remodeling of the heart characterized by the formation of fibrotic tissue that disrupts normal electrical conduction \cite{BAO:Elm2015,BAO:Mar2014}. Catheter ablation, a widely adopted AF treatment, seeks to isolate the abnormal electrical signals. However, accurate fibrosis detection and quantification can help to estimate the success of ablation strategies. Despite advancements in ablation techniques, recurrence rates of AF remain high, sometimes exceeding 40\% within 18 months following the procedure \cite{BAO:Ver2015}. These persistent challenges highlight the need for improved diagnostic tools capable of accurately assessing fibrosis patterns to enhance treatment selection and reduce AF recurrence.

Late Gadolinium Enhancement (LGE) MRI is widely utilized for detecting myocardial fibrosis and scarring, playing a crucial role in generating patient-specific models to plan ablation procedures \cite{BAO:Oak2009,BAO:Cai2021}. However, the diagnostic utility of LGE MRI is highly dependent upon image quality. Patient-related issues, such as movement and inconsistent breathing, can introduce motion artifacts. Additionally, suboptimal pulse sequence parameters and inadequate contrast administration may further degrade image clarity \cite{noisylge1,noisylge2,noisylge3}. Manual evaluation of image quality is labor-intensive and challenging to integrate into fast-paced clinical workflows. While deep learning offers a pathway to automate image quality assessment (IQA), traditional supervised approaches require exhaustive pixel-level or slice-level artifact annotations, which are costly and impractical to obtain for large-scale 3D LGE MRI datasets. To address this annotation bottleneck, Multiple Instance Learning (MIL) \cite{abmil} has emerged as a promising framework for learning under weak supervision. Specifically, the recent Sultan et al. \cite{sultan2024hamil} framework established the state-of-the-art for LGE-MRI quality assessment by employing a hierarchical architecture to predict volume-level diagnostic utility without explicit slice-level annotations.

However, standard MIL approaches, including recent hierarchical models, suffer from an inherent feature entanglement problem \cite{sultan2024hamil,abmil,dsmil,dtfdmil}. By mapping localized visual evidence to a single global feature vector, they operate as opaque black boxes. While this yields high accuracy, it obscures clinical reasoning. Furthermore, it leaves models susceptible to shortcut learning, where spurious visual cues artificially inflate performance instead of genuine anatomy. In clinical workflows, identifying that a scan is non-diagnostic is insufficient; technicians must know why: whether the failure is due to motion blur, inadequate contrast, or anatomical context. Concept Bottleneck Models (CBMs) \cite{cbm} attempt to address this by aligning latent dimensions with human-interpretable concepts. However, naive CBMs applied to medical images frequently suffer from spatial overlap and information leakage, where independent concepts rely on the same entangled visual textures \cite{inhocpaper,concept_leakage}.

To bridge this gap, we propose AC-MIL (Adversarial Concept Multiple Instance Learning), a generalized, interpretable MIL framework. While motivated by the critical need for transparent LGE-MRI quality assessment, AC-MIL is inherently domain-agnostic and applicable to any weakly-supervised medical imaging task where clinical concepts can be defined. To overcome the inherent limitations of standard architectures, we present a novel framework that successfully integrates concept bottlenecks into a weakly supervised MIL setting.
Our contributions are as follows:
\vspace{-1mm}
\begin{itemize}
    \item \textit{Disentangled Concept MIL Architecture:} We introduce a novel MIL formulation that explicitly decomposes the latent feature space into clinically meaningful concepts, transforming the opaque MIL pooling process into an interpretable clinical checklist.
    \item \textit{Adversarially-Regularized Completeness:} We introduce an unsupervised residual branch to model latent quality variations not described by the predefined concepts, utilizing an adversarial erasure mechanism to strictly prevent information leakage.
    \item \textit{Spatial Attention Diversity (SAD):} We propose a spatial diversity objective that penalizes overlap between the attention maps of distinct concepts to ensure localized and strictly interpretable feature extraction. 
\end{itemize}

\section{Method}
AC-MIL consists of two hierarchical tiers: (1) slice-level disentanglement into concept-specific subspaces and (2) volume-level aggregation under asymmetric gradient control. The framework is optimized via a multi-objective loss enforcing concept supervision, adversarial orthogonality, and spatial diversity.
\vspace{-5mm}
\subsection{Clinical Quality Concepts and Rating Scale}
The overall image quality for fibrosis assessment is evaluated on a $4$-point ordinal scale. We decompose this assessment into three predefined, weakly supervised volume-level concepts: \textit{Sharpness}, evaluating edge definition and chamber blurring; \textit{Myocardium Nulling}, assessing left ventricular intensity relative to the blood pool; and \textit{Aorta and Valve Enhancement}, evaluates the visibility and contrast of the aorta and valve leaflet walls. To capture the remaining latent variations critical for the final quality assessment, we introduce a fourth \textit{Unsupervised Residual} concept.
\vspace{-3mm}
\subsection{Problem Formulation and Architecture}
We represent each 3D MRI volume as a nested hierarchy of pseudo-bags. Utilizing segmentation masks, we restrict processing to the patient-specific number of axial slices ($M$) containing the Left Atrium. Let $\mathcal{V}$ be a bag of $M$ slices $\{S_1, \dots, S_M\}$, where $S_m \in \mathbb{R}^{H \times W}$. To accommodate the variable $M$ across patients during training, we randomly sample a fixed-size sub-bag of $N \le M$ slices. Each selected slice is further divided into a sub-bag of $K$ random patches $\{x_1, \dots, x_K\}$, with $x_k \in \mathbb{R}^{p \times p}$.
\begin{figure}[t]
\centering
\includegraphics[width=13cm]{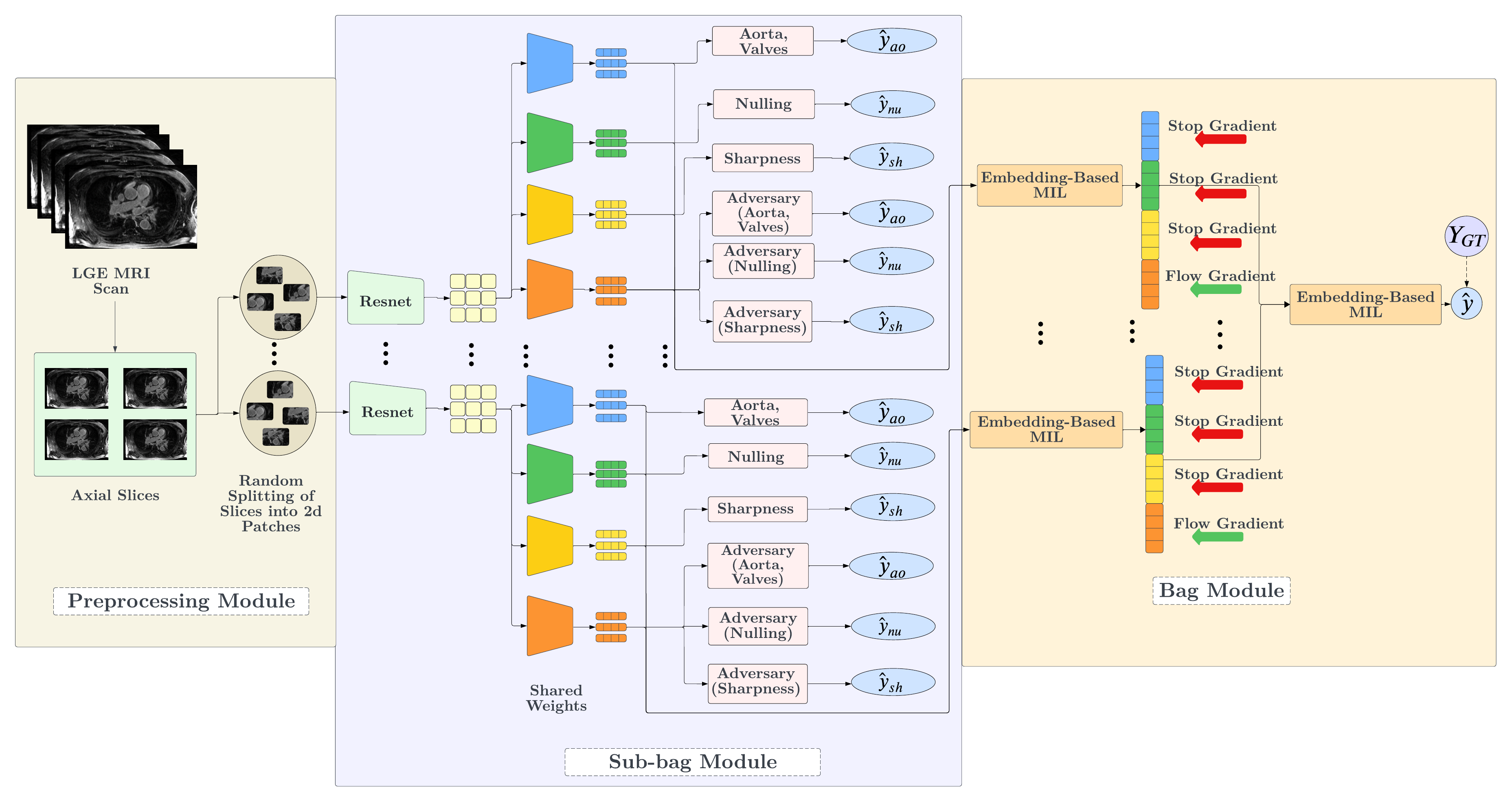}
\caption{\textbf{Architecture of the proposed AC-MIL framework.} 3D volumes are processed as a hierarchy of 2D slices and patches. Tier 1 (Sub-Bag Module) projects shared patch embeddings into three predefined clinical concept subspaces and one unsupervised branch (regularized by adversarial heads), using concept-specific attention for slice-level aggregation. Tier 2 (Bag Module) fuses these vectors via Attention-MIL to predict the final QA score.}
\label{arch}
\end{figure}
\vspace{-5mm}
\subsection{Tier 1: Disentangled Sub-Bag Module}
The first tier aims to decompose the entangled visual signals of a slice into our four distinct latent subspaces: Sharpness ($Z_{sh}$), Myocardium Nulling ($Z_{nu}$), Aorta/Valves ($Z_{ao}$), and the Unsupervised Residual ($Z_{un}$).

\vspace{0.05in}
\noindent \textbf{Concept-Specific Embedding.} Each patch $x_k$ is passed through a shared feature extractor $\Phi(\cdot)$ to obtain a feature embedding $h_k \in \mathbb{R}^{d}$, which is subsequently projected into $4$ equal-dimensional concept subspaces via dedicated 2-layer MLP heads $P^{(c)}$, yielding $h_k^{(c)} = P^{(c)}(h_k)\in{R}^{L}$ for $c \in \{sh, nu, ao, un\}$.

\vspace{0.05in}
\noindent \textbf{Concept-Specific Attention.} To aggregate patches, we employ independent attention mechanisms for each subspace, enabling selective focus on concept-specific regions. The slice-level vector is $Z^{(c)} = \sum_{k=1}^K \alpha_k^{(c)} h_k^{(c)}$, where attention weights $\alpha_k^{(c)}$ are computed via a concept-specific gated attention mechanism followed by softmax normalization across the $K$ patches.
\vspace{-2mm}
\subsection{Tier 2: Volume-Level Aggregation and Task Optimization}
The slice-level representations are fused into a comprehensive vector $V_m$. To prevent the downstream task loss from forcing predefined concepts to illicitly encode final task information, known as Concept Leakage \cite{concept_leakage}, we apply an asymmetric stop-gradient, $\text{sg}(\cdot)$, during fusion. The slice-level representations are concatenated as $V_m = [\text{sg}(Z_{sh}), \text{sg}(Z_{nu}), \text{sg}(Z_{ao}), Z_{un}]$, where the stop-gradient operator prevents task-level gradients from propagating into the predefined concept subspaces. Consequently, these branches are optimized solely via their respective supervised concept losses, while the residual branch remains the only pathway through which task gradients can flow. Note that gradients from the task loss still update the shared feature extractor through the residual pathway, but cannot directly modify the predefined concept projections.
A secondary Attention-MIL module then processes the sequence $V_{1 \dots M}$ to compute slice-level weights $\beta_m$ and aggregate them into a global volume representation $V_{vol} = \sum_{m=1}^M \beta_m V_m$.

A final classifier $F_{task}$ processes $V_{vol}$ to produce the volume-level Quality Assessment (QA) score. The primary objective, $\mathcal{L}_{task}$, optimizes this final prediction using the CORN loss \cite{corn_loss} against the global ground-truth volume label $y_{vol}$:
\begin{equation}
    \mathcal{L}_{task} = \mathcal{L}_{CORN}(F_{task}(V_{vol}), y_{vol})
\end{equation}
\vspace{-8mm}
\subsection{Learning via Multi-Objective Optimization}

To ensure the clinical relevance of these subspaces, we optimize a joint loss function:
\begin{equation}
    \mathcal{L}_{total} = \mathcal{L}_{task} + \lambda_1 \mathcal{L}_{CBM} + \lambda_2 \mathcal{L}_{adv} + \lambda_3 \mathcal{L}_{SAD}
\label{full_eqn}
\end{equation}
\noindent \textbf{Concept Supervision ($\mathcal{L}_{CBM}$):} We apply CORN losses \cite{corn_loss} to the predefined concept representations ${Z_{sh}, Z_{nu}, Z_{ao}}$. Due to the asymmetric stop-gradient, task-level gradients from $\mathcal{L}_{task}$ do not propagate into these branches. Consequently, their projections are optimized exclusively through $\mathcal{L}_{CBM}$, preventing direct task-driven concept leakage.

\vspace{0.05in}
\noindent \textbf{Adversarial Concept Erasure ($\mathcal{L}_{adv}$):} To prevent the unconstrained residual branch $Z_{un}$ from trivially relearning known semantics to minimize $\mathcal{L}_{task}$, we enforce feature orthogonality. Adversaries $D_{c}$ attempt to predict clinical concept grades from the unsupervised vector, formulated as a minimax game via a Gradient Reversal Layer (GRL)\cite{grl}:
\begin{equation}
    \mathcal{L}_{adv} = \sum_{c \in \{sh, nu, ao\}} \mathcal{L}_{CORN}(D_{c}(Z_{un}), y_c)
\end{equation}
During the backward pass, the GRL reverses the gradients by a factor of $-\lambda_2$ (i.e., $\frac{\partial GRL(x)}{\partial x} = -\lambda_2 \mathbf{I}$). This penalizes the feature extractor $\Phi(\cdot)$ if adversaries succeed, ensuring $Z_{un}$ remains statistically independent from predefined concepts.

\vspace{0.05in}
\noindent \textbf{Spatial Attention Diversity ($\mathcal{L}_{SAD}$):} We introduce a diversity constraint to prevent the model from collapsing multiple independent concepts onto the same anatomical patches. Crucially, global artifacts such as Sharpness affect the entire image and are therefore excluded from this penalty. We restrict the constraint to localized concepts by defining the set $\mathcal{C}_{loc} = \{nu, ao, un\}$. We then minimize the cosine similarity between the spatial attention maps of these specific pairs:
\begin{equation}
    \mathcal{L}_{SAD} = \sum_{\substack{c_i, c_j \in \mathcal{C}_{loc} \\ c_i \neq c_j}} \frac{\langle \alpha^{(c_i)}, \alpha^{(c_j)} \rangle}{\|\alpha^{(c_i)}\|_2 \|\alpha^{(c_j)}\|_2}
\end{equation}
By minimizing this overlap, we explicitly force the \textit{Aorta} head to attend to the vessel, the \textit{Nulling} head to focus on the myocardium, and the \textit{Unsupervised} head to capture quality information, guaranteeing spatial disentanglement.

\section{Experiments \& Results}
\subsection{Dataset}
We evaluated our framework on $775$ LGE MRI scans (resolution $0.625\times0.625\times2.5~mm^{3}$).
Data was partitioned via stratified 10-fold cross-validation, allocating an internal validation split for early stopping. Original expert quality ratings on a 1-5 scale were collapsed into a 4-point ordinal scale to mitigate class imbalance. We evaluate ordinal grading performance using Quadratic Weighted Kappa (QWK) and Average Mean Absolute Error (AMAE).
\vspace{-3mm}
\subsection{Implementation Details}
During training, we sample sub-bags of $N=8$ slices and $K=80$ patches ($64\times64$) per volume, whereas validation and testing employ dense sliding-window inference (50\% patch overlap) across all $M$ available slices. Models are optimized via AdamW \cite{adamw} (learning rate $1\times10^{-4}$, weight decay $1\times10^{-3}$) with a 100-epoch early stopping patience on validation QWK. Tuned on the validation set to balance ordinal grading accuracy with spatial disentanglement, the loss weights in Eq. \ref{full_eqn} are set to $\lambda_{CBM}=1.0$, $\lambda_{adv}=0.5$, and $\lambda_{SAD}=0.1$.
\vspace{-3mm}
\subsection{Qualitative Disentanglement \& Interpretability}
\label{section3_3}
\begin{figure}[!t]
    \centering
    \includegraphics[width=1\linewidth]{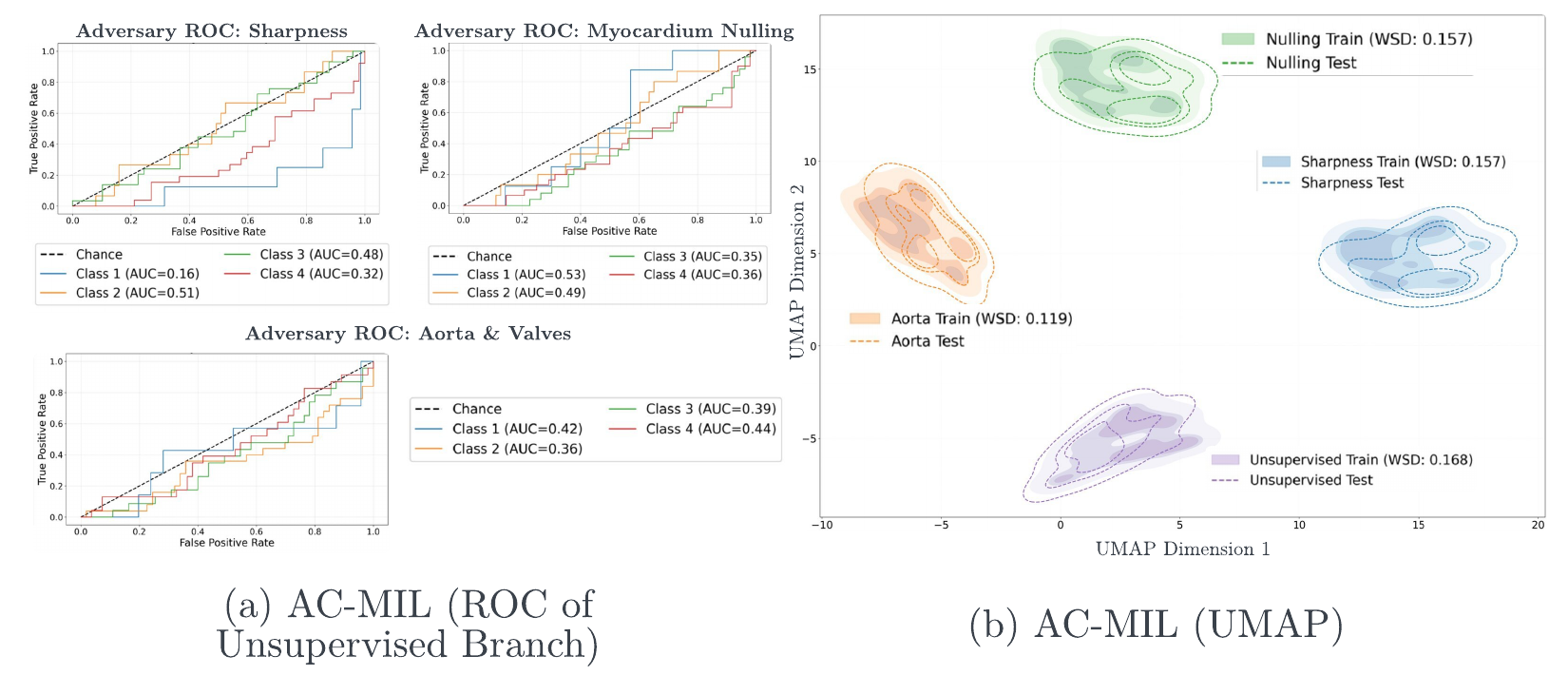}
    \caption{\textbf{Feature-space orthogonality and adversarial erasure.} (a) ROC curves of adversaries attempting to predict predefined concepts from the unsupervised branch. (b) UMAP projection of the learned feature spaces, with annotated Wasserstein Distances between training and test distributions.}
    \label{fig:umap}
\end{figure}
\begin{figure}[!t]
    \centering
    \includegraphics[width=1\linewidth]{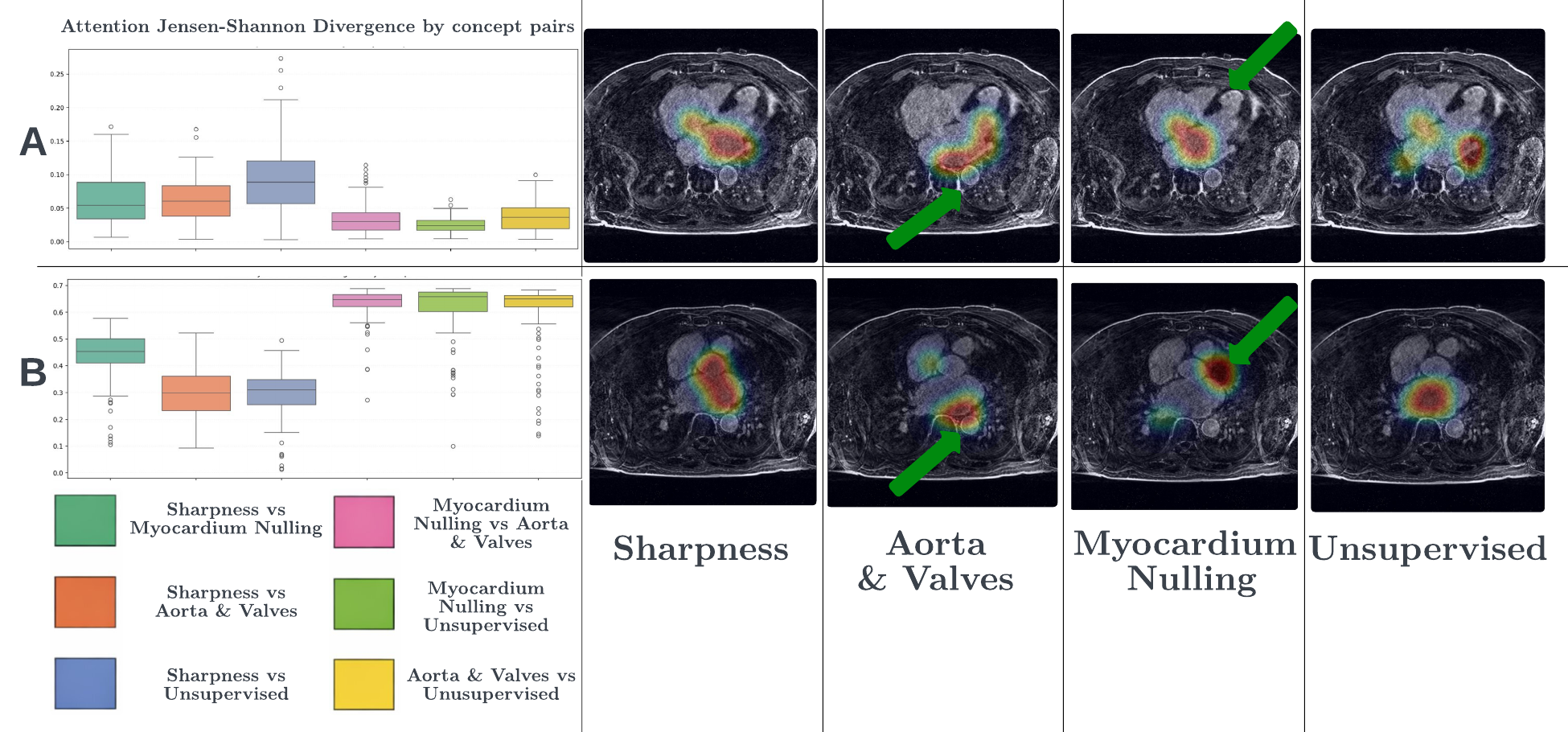}
    \caption{\textbf{Impact of the Spatial Attention Diversity ($\mathcal{L}_{SAD}$).} (A) Without $\mathcal{L}_{SAD}$, showing Jensen-Shannon Divergence (JSD) boxplots for concept pairs (left) and entangled spatial attention maps (right). (B) With $\mathcal{L}_{SAD}$ applied, showing increased JSD and attention precisely localized on target anatomies (green arrows).}
    \label{fig:jsd_attention}
\end{figure}
\noindent \textbf{Feature-Space Orthogonality and Erasure:} To validate the adversarial erasure mechanism, Figure \ref{fig:umap}a presents the ROC curves of adversaries attempting to predict predefined concepts from the unsupervised branch. The curves align closely with the random-chance diagonal, confirming that $\mathcal{L}_{adv}$ successfully purges known clinical signals from the residual space. Consequently, as visualized in the UMAP projection (Fig. \ref{fig:umap}b), the unsupervised representations do not entangle with existing features. Instead, they are forced to form a strictly distinct, non-overlapping fourth cluster alongside the predefined clinical concepts. 

\noindent \textbf{Spatial Disentanglement and Attention Diversity:} To evaluate feature localization, we computed the Jensen-Shannon Divergence between the spatial attention distributions of distinct concept pairs. Without $\mathcal{L}_{SAD}$, JSD is low (Fig. \ref{fig:jsd_attention}A), with Aorta and Nulling attention collapsing onto entangled, overlapping regions. This overlap allows the network to artificially inflate performance via shortcut learning, exploiting a single proxy variable rather than evaluating distinct clinical criteria. Applying $\mathcal{L}_{SAD}$ forces spatial specialization (Fig. \ref{fig:jsd_attention}B): predefined concepts decouple to target distinct anatomies (the aortic vessel and LV myocardium), while the unsupervised branch diverges to explore novel latent variations.

\begin{figure}[!h]
    \centering
    \includegraphics[width=1\linewidth]{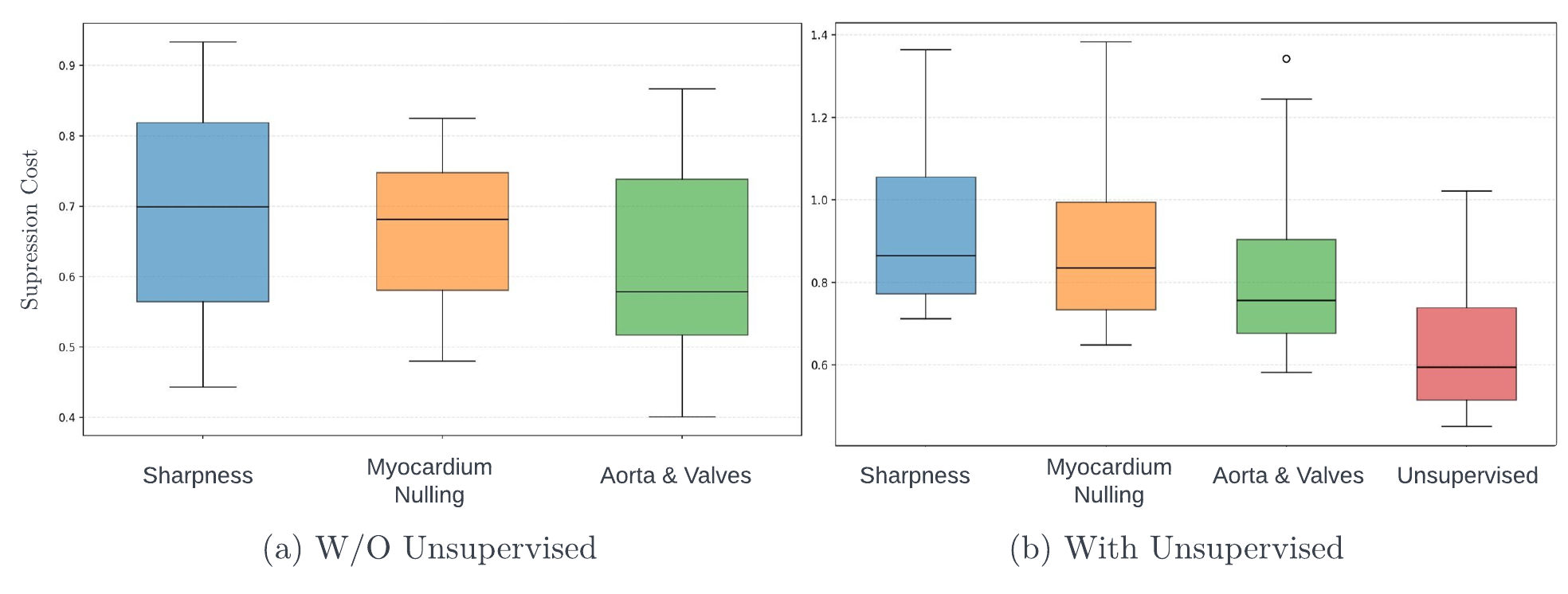}
    \caption{\textbf{Feature importance via adversarial attack.} Box plots display the minimum $L_2$ perturbation cost required to degrade a high-quality scan for models trained (a) without and (b) with the unsupervised residual branch.}
    \label{fig:feature_importance}
\end{figure}
\noindent \textbf{Clinical Relevance via Adversarial Attack:} Finally, we evaluated feature importance by calculating the minimum perturbation cost required to degrade high-quality scans via adversarial attacks\cite{adversarial} (Fig 4). While the baseline model distributes sensitivity across predefined concepts, the full AC-MIL framework is sensitive to the unsupervised residual. This sensitivity is clinically justified because while reference structures like the aorta and myocardium establish a baseline, they cannot account for every unique failure mode. The unsupervised branch thus acts as a specialized safety net, ensuring the final assessment accounts for complex anomalies that a standard clinical checklist would otherwise miss.
\vspace{-3mm}
\subsection{Quantitative Results and Ablation Study}
\begin{table}[!h]
\centering
\begin{tabular}{lccc|cc}
\toprule
\textbf{Method} & \textbf{CBM} & \textbf{SAD} & \textbf{Adv} & \textbf{QWK $\uparrow$} & \textbf{AMAE $\downarrow$} \\ 
\midrule
ABMIL\cite{abmil}        &             &              &              &  $0.60 \pm 0.03 [0.56, 0.64]$                      &     $0.59 \pm 0.01 [0.56, 0.63]$                       \\ 
HAMIL-QA\cite{sultan2024hamil}        &              &              &              & $0.63 \pm 0.01 [0.60, 0.65]$  & $0.57 \pm 0.02 [0.54, 0.61]$                           \\ 
AC-MIL                 & \checkmark   &              &              & $0.65 \pm 0.01[0.63,0.67]$                 & $0.56 \pm 0.01[0.53, 0.58]$                        \\ 
AC-MIL                & \checkmark   & \checkmark   &              &  $0.64 \pm 0.02 [0.61, 0.67]$             &  $0.57\pm0.02[0.53,0.61]$                          \\ 
AC-MIL               & \checkmark   &              & \checkmark   & $\mathbf{0.68\pm 0.01 [0.65,0.70]}$  &      $\mathbf{0.53 \pm 0.01 [0.50, 0.56]}$                      \\ 
\textbf{AC-MIL(Full)} & \checkmark   & \checkmark   & \checkmark   & $\mathcolor{blue}{\mathbf{0.66\pm0.01[0.64,0.67]}}$        &   $\mathcolor{blue}{\mathbf{0.55\pm0.01[0.52,0.57]}}$                         \\ 
\bottomrule
\end{tabular}
\caption{Ablation study on the components of our proposed framework on $10$ folds with 95\% confidence interval.}
\label{tab:ablation_checkmarks}
\end{table}
We benchmarked AC-MIL against SOTA methods \cite{abmil,sultan2024hamil} and evaluated our components via an ablation study (Table \ref{tab:ablation_checkmarks}). Our baseline AC-MIL variant establishes strong performance. However, as visualized in Fig. \ref{fig:jsd_attention}, this peak score relies heavily on the aforementioned spatial entanglement. Applying $\mathcal{L}_{SAD}$ corrects this overlap. The resulting marginal drop in the full model is a necessary trade-off to guarantee the network's reasoning aligns with true radiological anatomy.
\vspace{-5mm}
\section{Conclusion}
In this paper, we introduced AC-MIL, a novel weakly supervised framework for the interpretable quality assessment of atrial LGE-MRI scans. By preventing information leakage and penalizing shortcut learning, our approach successfully disentangles global quality predictions into human-interpretable radiological concepts. Extensive experiments demonstrate that by explicitly penalizing shortcut learning, AC-MIL guarantees that the network's reasoning aligns with true radiological anatomy, establishing the clinical trust necessary for automated quality assessment.

\section*{Acknowledgements} 
This work was supported by the National Institutes of Health under grant number NHLBI-R01HL162353. The content is solely the responsibility of the authors and does not necessarily represent the official views of the National Institutes of Health.
%

%
%
%
%

\bibliographystyle{splncs04.bst}
\bibliography{ac_mil.bib}

\end{document}